\documentclass[conference]{IEEEtran}
\IEEEoverridecommandlockouts
\usepackage{cite}
\usepackage{amsmath,amssymb,amsfonts}
\usepackage{graphicx}
\usepackage{textcomp}
\usepackage{xcolor}
\usepackage{float}
\usepackage[colorlinks=true, allcolors=black]{hyperref}
\usepackage{cite}
\usepackage{amsmath,amssymb,amsfonts}
\usepackage{graphicx}
\usepackage{textcomp}
\usepackage{xcolor}
\usepackage{subcaption}
\usepackage[english]{babel} 
\usepackage{graphicx}
\usepackage{subcaption}
\usepackage{graphicx} 
\graphicspath{ {./ITSC_FIG/} }
\usepackage[rightcaption]{sidecap}
\usepackage{graphics} 
\usepackage{epsfig} 
\usepackage{mathptmx} 
\usepackage{times} 
\usepackage{amsmath} 
\usepackage{amssymb}  
\usepackage[utf8]{inputenc}
\usepackage{mathrsfs}
\usepackage{booktabs}
\usepackage{amssymb }
\usepackage{amsfonts}
\usepackage{mathrsfs} 
\usepackage{comment}
\usepackage{titlesec}
\usepackage{amsmath,amsfonts,amssymb}
\usepackage{algorithm2e}
\usepackage[noend]{algpseudocode}
\usepackage{multirow}
\usepackage{longtable}
\usepackage{hyperref}
\usepackage{caption}
\def\BibTeX{{\rm B\kern-.05em{\sc i\kern-.025em b}\kern-.08em
    T\kern-.1667em\lower.7ex\hbox{E}\kern-.125emX}}

\makeatletter
\newcommand\subsubsubsection{\@startsection{paragraph}{4}{\z@}{-1ex\@plus -0.5ex \@minus -.25ex}{1.ex \@plus .25ex}{\normalfont\normalsize\bfseries}}
\newcommand\subsubsubsubsection{\@startsection{subparagraph}{5}{\z@}{-1ex\@plus -0.5ex \@minus -.25ex}{1.ex \@plus .25ex}{\normalfont\normalsize\bfseries}}

\makeatother

\def\BibTeX{{\rm B\kern-.05em{\sc i\kern-.025em b}\kern-.08em
    T\kern-.1667em\lower.7ex\hbox{E}\kern-.125emX}}
\begin{document}

\title{Toward a Holistic Multi-Criteria Trajectory Evaluation Framework for Autonomous Driving in Mixed Traffic Environment\\
\thanks{*The authors acknowledge the infrastructure and the support of the \textbf{SCARLET} team of \textbf{Vedecom institute}.
Special thanks to Benoît Lusetti and Alexis Warsemann for their involvement in the implementation and experimentation.}
}

\author{\IEEEauthorblockN{1\textsuperscript{st} Nouhed Naidja*}
\IEEEauthorblockA{\textit{CentraleSupelec, Laboratoire des signaux et systemes (L2S)} \\
Institut VEDECOM, Versailles, France \\
nihed.naidja@vedecom.fr, \quad nouhed.naidja@centralesupelec.fr}
\and
\IEEEauthorblockN{2\textsuperscript{nd} Stéphane Font}
\IEEEauthorblockA{\textit{CentraleSupelec, Laboratoire des signaux et systemes (L2S)} \\
Gif-sur-Yvette, France \\
stephane.font@centralesupelec.fr}

\and
\IEEEauthorblockN{ \hspace{3 cm} 3\textsuperscript{rd}  Marc Revilloud}
\IEEEauthorblockA{\hspace{3 cm} \textit{Dotflow} \\
\hspace{3 cm} Paris, France  \\
\hspace{3 cm} marc.revilloud@dotflow.fr}

\and 
\IEEEauthorblockN{ \hspace{2 cm} 4\textsuperscript{th} Guillaume Sandou}
\IEEEauthorblockA{\hspace{2 cm} \textit{CentraleSupelec, Laboratoire des signaux et systemes (L2S)} \\
\hspace{2 cm} Gif-sur-Yvette, France  \\
\hspace{2 cm} guillaume.sandou@centralesupelec.fr}
}
\maketitle
\begin{abstract}
This paper presents a unified framework for the evaluation and optimization of autonomous vehicle trajectories, integrating formal safety, comfort, and efficiency criteria. An innovative geometric indicator, based on the analysis of safety zones using adaptive ellipses, is used to accurately quantify collision risks. Our method applies the Shoelace formula to compute the intersection area in the case of misaligned and time-varying configurations. 
Comfort is modeled using indicators centered on longitudinal and lateral jerk, while efficiency is assessed by overall travel time. These criteria are aggregated into a comprehensive objective function solved using a PSO-based algorithm. The approach was successfully validated under real traffic conditions via experiments conducted in an urban intersection involving an autonomous vehicle interacting with a human-operated vehicle, and in simulation using data recorded from human driving in real traffic.
\end{abstract}


\section{Introduction}
Current research on autonomous vehicles and intelligent transport systems underlines the necessity for advanced decision-making frameworks that effectively manage multiple objectives.
Among these objectives, safety retains the highest priority, requiring the vehicles to not only avoid collisions, but also to comply with traffic rules as well as exhibit a predictable behavior in complex urban environments.
While safety is paramount, it is also essential to maintain the system's efficiency by optimizing traffic flows, minimizing delays, and reducing congestion, especially as transport infrastructures become increasingly interconnected.
In addition, passenger comfort, which encompasses not only the physical smoothness of maneuvers but also the psychological confidence of users in the system, has emerged as an important consideration influencing user acceptance and the deployment of autonomous driving technologies.
In light of the above, it is clear that balancing safety, efficiency, and comfort is not just a conceptual ideal but rather a requirement that shapes autonomous vehicle decision-making frameworks. 
This paper is structured around the formulation of criteria designed to evaluate the generated trajectory resulting from our approach, previously presented in our former work \cite{naidja2023interactive}. First, we focus on modeling safety-related metrics. Then, we formulate a comfort indicator by introducing jerk-based metrics. Efficiency evaluation will complete the two preceding metrics from the perspective of a multi-criteria evaluation. Once these criteria are designed, we proceed to a mathematical construction of the global objective function that will guide trajectory selection. We conclude with a synthesis of the proposed modeling framework. 

\section{RELATED WORK}
Achieving the right trade-off between safety, efficiency, and comfort in autonomous vehicles raises significant challenges, as these objectives may conflict in the context of real-time navigation and control.
For example, emphasizing safety by acting cautiously may increase delays, whereas performing dynamic maneuvers to optimize efficiency can compromise passenger comfort.
To address these trade-offs, recent approaches have explored how to reconcile these competing objectives by adopting unified frameworks that weigh these criteria.
In this context, the paper \cite{chen2023safe} presents a speed control approach using a Deep Deterministic Policy Gradient (DDPG) that trains an Autonomous Vehicle (AV) agent for optimal longitudinal acceleration during car following on rough pavements. 
The agent is penalized for near collisions and excessive jerk or acceleration, while rewarded for maintaining appropriate time headways. Simulation results show that the DDPG-based model exhibits significant improvements compared to both MPC-based (Model Predictive Control) and Adaptive Cruise Control (ACC) systems. Nevertheless, the reward function's structure, relying heavily on threshold-based penalties, might lead to abrupt driving maneuvers.
Exploring smoother penalty functions that gradually increase with deviation from desired behavior could improve the driving experience.
This issue is addressed by the authors in \cite{schildbach2016collision}. Instead of relying on precomputed assumptions about safety margins, the authors use an affine disturbance feedback mechanism that adjusts the control inputs of the ego vehicle based on the observed movements of the surrounding vehicles. This framework allows the system to react more smoothly to the real-time behavior of other vehicles, preventing sudden and jerky reactions. 

In light of the state of the art, this paper presents formal criteria for quantifying safety, efficiency, and passenger comfort. These criteria will be aggregated to construct agents' objective functions, which will subsequently be used to formulate a global optimization problem that allows for the simultaneous consideration of efficiency, comfort, and safety. Regardless of the method used to generate a trajectory, the criteria developed below provide elementary components that can be explored in a variety of applications.
They can be analyzed separately to assess the quality of a known trajectory. Also, they can be combined to create macro criteria for rating the quality of a trajectory in terms of both efficiency and comfort.
\section{Formalization of Safety, Comfort, and Efficiency Criteria for Trajectory Evaluation}
\label{sec:rec-car}

\subsection{Geometric Approach for Dynamic Safety Assessment}
To effectively quantify safety in autonomous driving scenarios, researchers often introduce the notion of Safety Zones surrounding the vehicle. These zones, defined as spatial boundaries or dynamic regions of permissible motion, represent the minimum required space needed to maintain safe maneuvers. Common geometrical representations of safety zones include circles, ellipses, rectangles, and other polygonal or parametric shapes \cite{nilsson2017trajectory}. 
The simplest approach is to model a safety margin as a circular area \cite{cai2021game}. This model is computationally efficient and easy to integrate into optimization problems.
However, circular safety zones do not account for the shape or directional properties of the vehicle. As a result, they require the use of a large radius that covers all possible situations. Rectangular and polygonal safety zones \cite{nilsson2017trajectory} offer a more precise alternative to circular zones. However, this precision comes with challenges: computing polygonal intersections for collision detection is computationally intensive \cite{wang2016obstacle}. 
Elliptical safety zones provide a shape that closely aligns with the vehicle's operational profile while balancing the simplicity of circular zones and the precision of polygonal approximations \cite{wang2023driving}. Their axes can be adjusted to reflect changes in braking distance or lateral clearance, allowing them to adapt smoothly to vehicle dynamics.
In this paper, we build upon the advantages of elliptical safety zones and leverage their ability to capture asymmetric safety margins to formulate a geometric collision avoidance framework.

\subsection{Geometric Approach for Dynamic Safety Assessment}

\begin{figure}[H]
    \centering
     \includegraphics[width=0.6\linewidth]{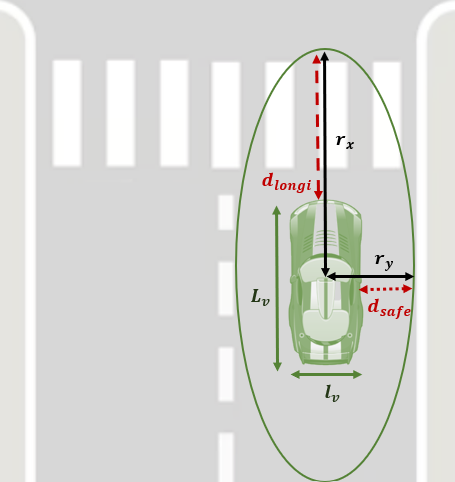} 
   \caption{Adaptive Elliptical Safety Zone}
    \label{fig: adaptive Safety Zone}
\end{figure}

We extend the concept of elliptical safety zones to provide a dynamic and adaptive representation of the space each vehicle is expected to occupy, reflecting both geometric and dynamic characteristics.
We propose a geometric safety criterion based on the spatial relationship between two ellipses, each representing the safety zones around two interacting vehicles.
Our goal is to guarantee that these safety zones remain separated, mitigating the risk of collisions in dynamic environments.
The following section builds upon the work presented in our previous paper \cite{naidja2024gtp}.
Unlike static geometric shapes, our safety zones are continuously adjusted to account for both geometric and dynamic considerations, including the vehicle's shape, velocity, position, and orientation. This dynamic adaptation allows the ellipses to serve as predictive zones for potential collisions, rather than static spatial boundaries.
Specifically, the parameters of the ellipses are computed using the vehicle’s length \(L_{v}\) and width \(l_{v}\). 
The ellipse's major axis, \(r_x\), is continuously adjusted according to the vehicle's velocity \({v}(t)\).
To further enhance this framework, we incorporate the Time to Collision (TTC) threshold as a risk assessment indicator. Further explanations on the computation and the use of this indicator are given \cite{zhu2020safe}.
 The TTC represents the time required for a vehicle to reach a potential collision if it maintains its current speed and trajectory, and under the assumption that the vehicle's movement is quasi-aligned with its longitudinal axis. Building upon the TTC, we compute a longitudinal safety margin as \(d_{longi}= \text{{TTC}}_{th}\cdot\textit{v}(t)\), with $TTC_{th}$ the time-to-collision threshold.
This metric is used to adjust the longitudinal semi-axis of the ellipse \(r_x\) to reflect dynamic conditions, ensuring sufficient separation from other vehicles.
On the other hand, the lateral safety margin \({d}_{safe}\) shown in Figure \eqref{fig: adaptive Safety Zone} contributes to the minor semi-axis \(r_y\).
In practice, \(d_{safe}\) serves two purposes. First, it ensures that the safety zone remains within the road boundaries, preventing the vehicle from encroaching onto adjacent lanes. Second, it promotes a centered driving position by balancing the lateral margin symmetrically around the vehicle within the lane.
To achieve this dynamic behavior, \(r_x\) and \(r_y\), the major and minor semi-axes of the ellipse, are dynamically updated according to equation \eqref{equ: adaptiveElli}: 

\begin{align}
 \normalsize
\left\{\begin{matrix}
r_x=\frac{1}{2} {L}_{v} + d_{longi}\\
r_y=\frac{1}{2} {l}_{v} + d_{safe}\end{matrix}\right.
\label{equ: adaptiveElli}
\end{align}

Here \(r_x\) evolves based on the longitudinal safety margin \(\boldmath d_{longi}\), as illustrated in Figure \eqref{fig: adaptive Safety Zone}, reflecting the risk of a forward collision.
The minor semi-axis \(r_y\) remains primarily influenced by the vehicle’s lateral geometry, including the vehicle width \(l_{v}\) and a lateral safety margin \(d_{safe}\).

In the present study, we extend this framework by focusing on the exploitation of these ellipses to assess the safety of interactions.

\subsection{Collision Risk Analysis: Minimum Distance and Overlap Metrics}

\begin{figure}[htp]
    \centering
     \includegraphics[width=0.7\linewidth]{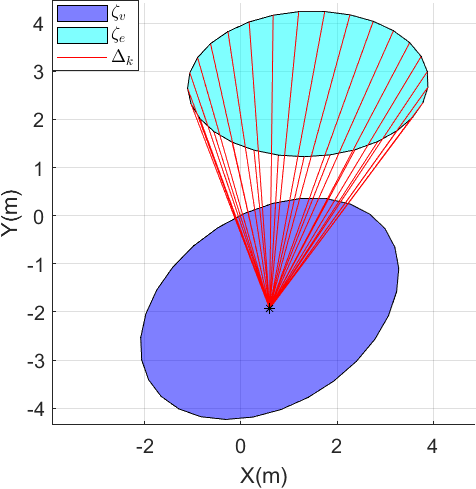} 
    \caption{Relative Position Vectors \( \overrightarrow{\Delta_k} \)}
    \label{fig: Delta}
\end{figure}

Consider the two ellipses in the figure \eqref{fig: Delta}. The purple ellipse \( \boldsymbol \zeta_{v} \), centred at \( O(x_c, y_c)\), representing the safety zone of the ego vehicle \(\boldsymbol {v}\), is defined by its semi-axes \({r_x }\) and \({r_y}\), and rotation angle \( \boldsymbol \theta\)  relative to the axes of the global frame \( {R_0} \). The bleu ellipse \( \boldsymbol \zeta_{o} \), centred at  \( \bar O( \bar x_c, \bar y_c)\) defines the opponent vehicle's safety zone, with points on its boundary denoted as \( \overline{m}_k ( \bar{x}_k, \bar{y}_k) \). 
We consider the relative position vectors \( \overrightarrow{\Delta_k} \) that captures the relative position between \( O(x_c, y_c)\), the center of the ellipse \(\zeta_{v} \), to each boundary point \( \overline{m}_k ( \bar{x}_k, \bar{y}_k)  \) of the opponent ellipse \(\zeta_{o} \) (see the figure \eqref{fig: Delta}),  where \( k\) denotes the sampled angles along the boundary of the ellipse \(\zeta_{o} \).
These vectors are initially expressed in \( {R_1} \) as: 
\begin{equation}
    \overrightarrow{O \overline{m}_k } = \overrightarrow{\Delta_k} \mid_{R_1} =
\begin{bmatrix}
\bar{x}_k - x_c   \\
\bar{y}_k - y_c 
\end{bmatrix},
 \quad k \in [0, 2\pi]
\label{equ:OM}
\end{equation} 
We express the vector $\overrightarrow{\Delta_k} $
\begin{equation}
\tilde{\Delta}_k = 
\begin{bmatrix}
\displaystyle \frac{\cos(\theta) (\bar{x}_k - x_c) + \sin(\theta) (\bar{y}_k - y_c)}{r_x} \\
\displaystyle \frac{-\sin(\theta) (\bar{x}_k - x_c) + \cos(\theta) (\bar{y}_k - y_c)}{r_y}
\end{bmatrix}
\label{equ:Delta_k}
\end{equation}

\subsection{Intersection Detection} 

We construct a geometric safety criterion to rigorously analyze the spatial relationship between the two ellipses representing the safety zones surrounding the vehicles. 

Our approach evaluates the proximity or the severity of overlap between these two ellipses by introducing two complementary measures based on their spatial configuration. We identify two possible scenarios: 
\begin{itemize}
    \item  \( \boldsymbol{\forall k\in[0, 2 \pi]}, ~ {\|\tilde{\Delta}_k \| > 1}\): \textbf{Non-Intersection Scenario \(\zeta_{v}  \bigcap \zeta_{o} = \emptyset\) }: the ellipses remain separated. We compute a separating distance, noted \(D\). This distance acts as an indicator of how far apart the safety zones are, ensuring that vehicles maintain a safe separation margin.
    
    \item \(\boldsymbol{\exists k\in[0, 2 \pi]}, ~ {\|\tilde{\Delta}_k \| \leq 1}\): \textbf{Intersection Scenario \(\zeta_{v}  \bigcap \zeta_{o} \neq  \emptyset\)}:
    If at least one norm  \( \|\tilde{\Delta}_k \|  \leq 1 \), the ellipses intersect, indicating a violation of the minimum safety margin. We compute the area of overlap \(A\) between the ellipses. This area serves as a measure of the severity of the intrusion between the safety zones.
\end{itemize}

This approach ensures a comprehensive evaluation of the spatial relationship between vehicles, whether their safety zones overlap or remain separated. 

\subsection{\textbf{Non-Intersection Scenario – Minimum Remaining Gap}}
\label{sec:Non_intersec}
The distance $d^2$ between these two ellipses $\zeta_v$ and $\zeta_o$ is defined as follows:
\begin{equation}
d^2 = \inf_{\substack{k \in [0, 2\pi] \\ \bar{k} \in [0, 2\pi]}} \left\| \begin{pmatrix} x(k) \\ y(k) \end{pmatrix} - \begin{pmatrix} \bar{x}(\bar{k}) \\ \bar{y}(\bar{k}) \end{pmatrix} \right\|^2
\label{equ:d_app_1}
\end{equation}

Alternatively, the analysis can be restricted to points on each ellipse that correspond to each other through radial projection from the respective centers. 
The radial projection corresponds to the mappings that assign to any point the unique point on the ellipse obtained by projection along the radius from its center.
For any point other than the center of the ellipse, the vector \( \tilde{\Delta}_k  \) is non-zero and can be normalized. The resulting unit-norm vector defines a point on the ellipse, corresponding to the radial projection of the original point onto the ellipse. We denote this point by  \( P_{\zeta_{v}(m_k)}\):
\begin{equation}
P_{\zeta_{v}(m_k)} = \begin{bmatrix}
x_c \\ y_c
\end{bmatrix}^{\top} + 
\begin{bmatrix}
\cos(\theta) & -\sin(\theta) \\
\sin(\theta) & \cos(\theta)
\end{bmatrix} \cdot \left( \frac{\tilde{\Delta}_k}{\|\tilde{\Delta}_k\|} \right)
\end{equation}
Similarly, the radial projection onto \(\zeta_{o}\) is defined in the same way and will be denoted by \( P_{\zeta_{o}}\).

Thus, the distance between the ellipses can be equivalently defined as:
\begin{equation}
d^2 = \inf_{\substack{k \in [0, 2\pi] \\ \bar{k} \in [0, 2\pi]}} \left\| P_{\zeta_v} \left( \begin{pmatrix} \bar{x}(\bar{k}) \\ \bar{y}(\bar{k}) \end{pmatrix} \right) - P_{\zeta_o} \left( \begin{pmatrix} x(k) \\ y(k) \end{pmatrix} \right) \right\|^2
\label{equ:d_app_2}
\end{equation}

In practice, angular sampling is used to represent the ellipses. Let: $\Omega_N = \left\{ \frac{2\pi n}{N} \,\middle|\, n = (0, \dots, N-1) \text{ samples} \right\}$.
The distance expressed previously in equation \eqref{equ:d_app_2} is approximated by the following:
\begin{equation}
    d^2_{\text{app}} =\min_{\substack{k \in \Omega_N \\ \bar{k} \in \Omega_N}} \left\| P_{\zeta_v} \left( \begin{pmatrix} \bar{x}(\bar{k}) \\ \bar{y}(\bar{k}) \end{pmatrix} \right) - P_{\zeta_o} \left( \begin{pmatrix} x(k) \\ y(k) \end{pmatrix} \right) \right\|^2
    \label{equ:d_approxi}
    \end{equation}

By construction, this approximation overestimates the real distance, as it relies on a finite subset of perimeter points and is strictly positive. Hence, $d \in [0, d_{\text{app2}}]$.
Since the computation of $d_{\text{app}}$ has a complexity of order $O(N^2)$ and, in fine, would be integrated into an optimization framework, we need to define a distance indicator that remains representative while reducing the computational cost.
Thus we define the associated points $N_v \in \zeta_v$ and $N_o \in \zeta_o$ as:
\begin{align}
N_v &= \arg\min_{\bar{k} \in \Omega_N} \left\| P_{\zeta_v} \left( \begin{pmatrix} \bar{x}(\bar{k}) \\ \bar{y}(\bar{k}) \end{pmatrix} \right) - \begin{pmatrix} \bar{x}(\bar{k}) \\ \bar{y}(\bar{k}) \end{pmatrix} \right\|^2 \\
N_o &= \arg\min_{k \in \Omega_N} \left\| \begin{pmatrix} x(k) \\ y(k) \end{pmatrix} - P_{\zeta_o} \left( \begin{pmatrix} x(k) \\ y(k) \end{pmatrix} \right) \right\|^2
\end{align}

\begin{figure}[htp]
    \centering
     \includegraphics[width=0.75\linewidth]{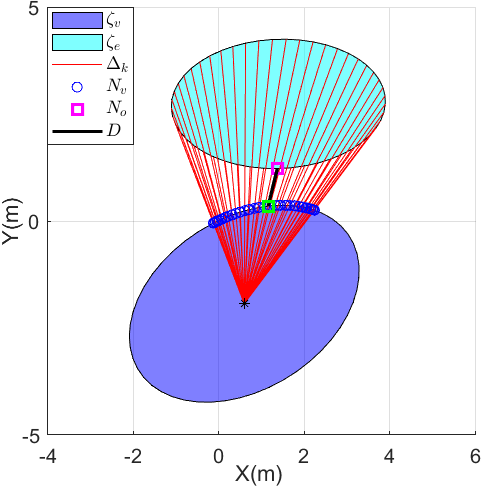} 
    \caption{Distance Criterion: Minimum Remaining Gap}
    \label{fig: Delta_nearest}
\end{figure}
The overall indicator $D$, illustrated in Figure \eqref{fig: Delta_nearest}, is then defined as the Euclidean distance between these two points:
\begin{equation}
D = \left\| N_v - N_o \right\|
\end{equation}

In the non-intersection scenario, the indicator D provides a robust measure of the minimum gap between the safety zones, ensuring sufficient safety margins and adherence to road boundaries. 
However, as dynamic traffic conditions evolve, there may arise situations where \(\tilde{\Delta}_k  \leq 1 \), indicating an overlap of the safety zones. 
Such cases represent potential safety violations, requiring a more detailed analysis.

\subsection{\textbf{Intersection Scenario – Overlap Quantification}}
In this section, our focus shifts to the characterization and quantification of the overlap area between the ellipses.
When two ellipses overlap, the resulting intersection area A presents a complex geometric shape, and in general, no simple closed-form formula exists to compute the exact area, especially when the ellipses are not aligned.

In this paper, we adopt a geometric method that represents the overlap as a convex polygon and leverages the \textbf{Shoelace Formula} (Gauss’s Area Formula) \cite{braden1986surveyor} for efficient area computation.
The shoelace formula is a rigorous method for computing the area of a convex polygon given the ordered coordinates of its vertices. If the boundary of a shape can be represented as a polygon, the shoelace formula provides a direct, closed-form expression for its area. 
Let \( P \subset \mathbb{R}^2 \) represent a simple polygon in the Euclidean plane, defined by its \( n \)-ordered vertices $V = \{(x_1, y_1), (x_2, y_2), \dots, (x_n, y_n)\}$, traversed in, counter-clockwise cyclic order, and assumed closed with \( (x_{n+1}, y_{n+1}) = (x_1, y_1)\). 
The area \( A(P) \) of the polygon \( P \) can be expressed as:
\begin{equation}
A(P) = \frac{1}{2} \left| \sum_{i=1}^{n} \left( x_i \cdot y_{i+1} - y_i \cdot x_{i+1} \right) \right|
\label{eq:polygon_area}
\end{equation}
Where \( (x_i, y_i) \in \mathbb{R} \) are the Cartesian coordinates of the \( i \)-th vertex.
However, the formula in equation \eqref{eq:polygon_area} does not apply to continuous curves such as circles or ellipses unless those curves have first been discretized or approximated by polygonal segments.

To use the shoelace formula to compute the intersection area of two safety zones, we apply a discretization step by approximating the shared region between the two ellipses with a sufficiently fine polygonal mesh. 

\begin{figure} [h!]
     \centering
      \hspace{-0.8cm}
     \begin{subfigure}[b]{0.28\textwidth}
         \centering
         \includegraphics[width=0.85\textwidth]{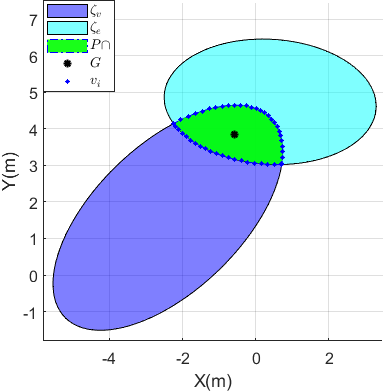}
         \caption{Discretization of the Intersection Area}
         \label{fig:Ellipse-Ellipse Intersection}
     \end{subfigure}
       \hspace{-0.85cm}
     \begin{subfigure}[b]{0.28\textwidth}
         \centering
         \includegraphics[width=0.85\textwidth]{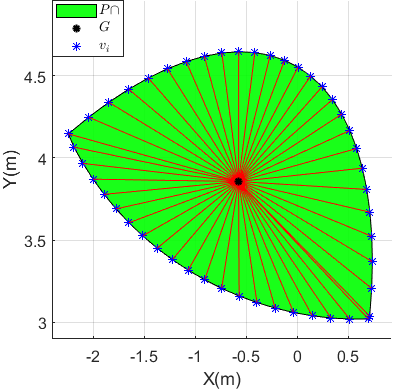}
         \caption{ \( P \cap \) Approximated by Union of Triangles}
         \label{fig:Approximated Polygone}
     \end{subfigure}
        \caption{Discretization and Polygonal Approximation of the Intersection of Two Ellipses}
        \label{fig:Configuration}
\end{figure}

First, we determine the subset of points from each ellipse that are contained within the other, as shown in Figure \eqref{fig:Ellipse-Ellipse Intersection}.
As a result, we obtain a collection of points ~ $v_i = (x_i, y_i)$ that define the perimeter of the intersection region. Since these points are not ordered, it is necessary to sort them in a counter-clockwise direction. 
Second, we compute the centroid \( \mathbf{G} = (x_G, y_G) \) of the polygon  as the arithmetic mean of its vertices: $x_G = \frac{1}{n} \sum_{i=1}^{n} x_i, \quad
y_G = \frac{1}{n} \sum_{i=1}^{n} y_i,$, where \( n \) is the total number of vertices, and \( (x_i, y_i) \) are the Cartesian coordinates of the \( i \)-th vertex.
This centroid serves as a reference point for sorting the vertices and ensures a consistent orientation.
To achieve a counter-clockwise ordering, the angle \( \angle G v_{i} \) between the centroid \( G \) and each vertex \(v_{i} (x_i, y_i)\) is calculated, and the vertices are sorted in ascending order of \(\angle G v_{i}\).
Once this set of vertices is arranged counter-clockwise,  they define the convex hull denoted \( P \cap \) that approximates the actual ellipse-ellipse intersection in the continuous domain. 
We consider the origin $G$ inside \( P \cap \) and connect it to each vertex  $v_{i}=(x_i, y_i)$, thereby decomposing \( P \cap \) into $n$ triangles, as illustrated by the figure \eqref{fig:Approximated Polygone}.

Consider a single triangle $\Delta(G, v_i, v_{i+1})$, with $G$ chosen as origin for simplicity, and $v_i = (x_i, y_i)$, $v_{i+1} = (x_{i+1}, y_{i+1})$. In vector notation, let: $\vec{Gv}_{i} = \begin{pmatrix} x_i \\ y_i \end{pmatrix}, \quad \vec{Gv}_{i+1} = \begin{pmatrix} x_{i+1} \\ y_{i+1} \end{pmatrix}$

The signed (or oriented) area of this triangle is expressed by the following:
\begin{equation}
\frac{1}{2}   \left| \vec{Gv}_{i}\wedge \vec{Gv}_{i+1} \right|= \frac{1}{2} \det \begin{pmatrix} x_i & x_{i+1} \\ y_{i} & y_{i+1} \end{pmatrix} = \frac{1}{2} \left| x_i \cdot y_{i+1} - y_i \cdot x_{i+1}\right|
\label{eq:Sum_signed}
\end{equation}

Summing from $i = 1$ to $n$ gives:
\begin{equation}
\sum_{i=1}^n \frac{1}{2} (x_i \cdot y_{i+1} - y_i \cdot x_{i+1} ) = \frac{1}{2} \sum_{i=1}^n (x_i \cdot y_{i+1} - y_i \cdot x_{i+1} )
\label{eq:Sum}
\end{equation}

To recover the usual “unsigned” area, we take the absolute value as given below:
\begin{equation}
A = \frac{1}{2} \left| \sum_{i=1}^n (x_i \cdot y_{i+1} - x_{i+1} \cdot y_i) \right|
\label{eq:InsugnedSum}
\end{equation}
Hence, we arrive at the Shoelace formula presented in the Equation \eqref{eq:polygon_area}.

We have thus identified two critical metrics to evaluate the spatial relationship between the safety zones of two vehicles.
These metrics provide a quantitative basis for a comprehensive evaluation of safety in dynamic environments. 
In the next section, we explore how they can be integrated to construct a safety criterion that can guide collision avoidance strategies.


\subsection{Toward a Safety Criterion for Trajectory Optimization}
\label{sec:safety}

This section introduces a methodology to integrate geometric metrics, namely the distance indicator \( D \) and the intersection area \( A(P \cap) \), into a unified safety criterion, denoted by \( C_{\text{safe}} \). 
The approach outlined here aims to ensure that the optimization process inherently favors collision-free and sufficiently spaced trajectories.
First, we define the intersection metric \(\text{Int}(t)\) at a given time step \(t\). Formally:
\begin{equation}
\text{Int}(t) =
\begin{cases}
A(P \cap), & \text{if } \zeta_v \cap \zeta_e \neq \emptyset \quad (\text{Overlap}), \\
-D, & \text{if } \zeta_v \cap \zeta_e = \emptyset \quad (\text{Non-overlap}).
\end{cases}
\label{eq:interaction_metric}
\end{equation}
In Equation~\eqref{eq:interaction_metric}, larger \( A(P \cap) \) corresponds to a more severe safety margin violation.
Conversely, if there is no overlap (\(\zeta_v \cap \zeta_o = \emptyset\)), the metric \(\text{Int}(t)\) returns \(-D\), where \( D \) denotes the minimum distance between the two ellipses. Smaller (more negative) values of \(-D\) signify larger and, thus, safer separation distances.
The intersection metric \(\text{Int}(t)\) is evaluated at each discrete time step \(i\), thereby enabling a time-dependent representation of safety. 

Building upon the intersection metric, defined in Equation~\eqref{eq:interaction_metric}, we now introduce the safety criterion \(C_{\text{safe}}\), expressed as:
\begin{equation}
C_{\text{safe}} = \beta \left( \max_t (\text{Int}(t)) \right)
\label{eq:safe_cost}
\end{equation}

In Equation~\eqref{eq:safe_cost}, the term \(\max_{t} (\text{Int}(t)) \) isolates the worst-case interaction among all time steps. 
Focusing on this maximum measure, the safety criterion highlights the time at which the most critical safety violation (the smallest gap or overlap) occurs, rather than averaging over multiple time steps. 

Moreover, the shaping function $\beta(x)$, introduced in the safety criterion  \(C_{\text{safe}}\) (Equation~\eqref{eq:safe_cost}), modulates how different spatial configurations contribute to the overall cost. It is defined as:
\begin{equation}
\beta(x) =
\begin{cases}
M\left(1 - \frac{px}{M\alpha}\right)^{-\alpha} & \text{if } x \leq 0 \\
M e^{\frac{p}{M}x} & \text{if } x > 0
\label{eq:beta_function}
\end{cases}
\end{equation}

\begin{figure}[h!]
    \centering
     \includegraphics[width=0.75\linewidth]{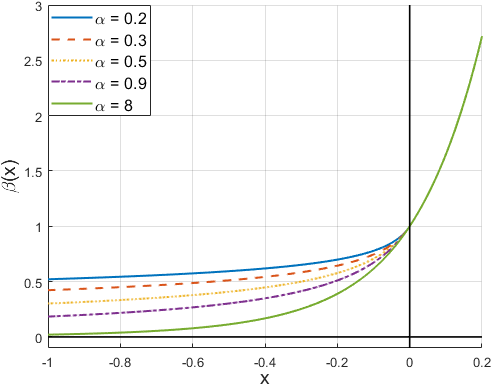} 
    \caption{An illustration of the function Shaping function $\beta(x)$: parameter values $M = 1$, $p = 5$, and $\alpha \in \{0.2, 0.3, 0.5, 0.9, 8\}$ One can observe that $\beta(x)$ converges more rapidly to zero as $x \to -\infty$ for larger values of $\alpha$.}
    \label{fig:Shaping_function}
\end{figure}

The function $\beta$ is continuous and differentiable, with the following key properties:
\begin{equation}
\lim_{x \to 0^-} \beta(x) = \lim_{x \to 0^+} \beta(x) = \beta(0) = M
\end{equation}
\begin{equation}
\lim_{x \to 0^-} \beta'(x) = \lim_{x \to 0^+} \beta'(x) = \beta'(0) = p
\end{equation}
\begin{equation}
\lim_{x \to -\infty} \beta(x) = 0, \quad \lim_{x \to +\infty} \beta(x) = +\infty
\end{equation}
Where $\alpha > 0$, $M > 0$, $p > 1$. 
The figure \eqref{fig:Shaping_function} illustrates how the parameter $\alpha$ influences the shaping function $\beta(x)$. All curves share the same value for $p$  and $M$, while $\alpha$ varies. 
The parameters $M$ (the value at $\beta(0)= M$), and p ($\beta'(x) =  p$)  govern the increase of the exponential function. 
We can distinguish two regions depending on the value of $x$: for positive values of $x$, corresponding to an overlap scenario, we can see that $\beta(x)$ grows exponentially, which reflects a strong penalization of overlapping configurations. A high $p/M$ ratio leads to significant penalties even for slight overlaps.
For negative values of $x$, which indicate non-overlapping configurations, the shape of $\beta(x)$ is governed by the parameter $\alpha$. It controls the decay rate of the function $\beta(x)$ as $x \to -\infty $, and thus how the criterion considers safety margins. For large values of $\alpha$, the function $\beta(x)$ flattens quickly once the distance is sufficient to avoid overlap, meaning further increases have minimal effect on the cost. 
In contrast, smaller values of $\alpha$ decrease the function gradually, thus encouraging larger safety margins even when no overlap occurs.
Overall, the proposed framework for \(C_{\text{safe}}\) aligns with the goal of systematically penalizing risky maneuvers. Consequently, an optimization algorithm leveraging this criterion will prioritize solutions where collisions are avoided and safety margins are maintained throughout the optimization process. 
However, safety alone is not sufficient to design an optimal driving strategy.
While avoiding collisions is the highest priority, an autonomous vehicle must also ensure a smooth and comfortable ride for passengers.
To address this aspect, the following section introduces a comfort criterion, which promotes smooth variations in acceleration and steering.

\subsection{Jerk-Centric Comfort Evaluation for Autonomous Driving }
\label{sec:Indicators_Confort}
In the context of autonomous driving, comfort addresses the qualitative dimension of AV navigation, focusing on ensuring a smooth ride. Metrics linked to the vehicle's dynamics, including acceleration and jerk, capture the passengers' comfort \cite{ 10155306}. 
The jerk quantifies the rate of change of acceleration over time. Excessive jerk values result in undesirable vehicle oscillations and increasing discomfort. Therefore, commonly used metrics to assess comfort in trajectory planning include jerk minimization.
In this section, we define the mathematical formulation of the comfort criterion by analyzing longitudinal jerk, denoted \( j_{long}\) and lateral jerk, denoted \( j_{lat}\), employing finite difference approximations for numerical computation.
\subsubsection{Comfort Indicators} 
\label{sec:Indicators_Confort}
Let $\beta(t)$ denote the angle representing the orientation of the vehicle’s velocity vector $V(t)$, with respect to a fixed reference orientation.
The velocity, acceleration, and jerk are expressed in this Frenet frame as follows: 
\begin{align}
V(t) &= v(t)\cdot T(\beta(t)) \\
a(t) &= \frac{d}{dt} \left[V(t)\right] 
= \dot{v}(t)\cdot T(\beta(t)) + v(t)\cdot\beta'(t)\cdot N(\beta(t)) \\[1.5ex]
j(t) &= \frac{d}{dt} \left[a(t)\right] \nonumber=\left[\ddot{v}(t) - v(t) \cdot (\beta'(t))^2\right] T(\beta(t)) \nonumber\\
&\quad + \left[2\dot{v}(t) \cdot \beta'(t) + v(t) \cdot \beta''(t)\right] N(\beta(t))
\end{align}

With $T(\beta(t))$, and $N(\beta(t))$ representing the unit tangent and normal vectors, respectively. Thus, we identify the components:

\textbf{\textendash Longitudinal jerk component:}
\begin{equation}
j_{\text{long}}(t) = \ddot{v}(t) - v(t) \cdot (\beta'(t))^2
\label{equ:long_appro}
\end{equation}

\textbf{\textendash Lateral jerk component:}
\begin{equation}
j_{\text{lat}}(t) = 2\dot{v}(t) \cdot \beta'(t) + v(t) \cdot \beta''(t)
\label{equ:lat_appro}
\end{equation}
When estimating the time derivative of kinematic variables from discrete data, the choice of the finite difference order is not trivial.
In this paper, we opted for a second-order yielding an approximation error of order $\mathcal{O}(\Delta t^2)$, which is generally acceptable for practical application. 
From the decomposition in Equations \eqref{equ:long_appro} and \eqref{equ:lat_appro}, we approximate:

\begin{equation}
j_{\text{long}}(t_i) \approx \frac{v_{i+1} - 2v_i + v_{i-1}}{\Delta t^2} - v_i \cdot \left( \frac{\beta_{i+1} - \beta_{i-1}}{2\Delta t} \right)^2
\label{eq:longi_jerk_final}
\end{equation}

\begin{equation}
j_{\text{lat}}(t_i) \approx \frac{v_{i+1} - v_{i-1}}{\Delta t} \cdot \frac{\beta_{i+1} - \beta_{i-1}}{2\Delta t} + v_i \cdot \frac{\beta_{i+1} - 2\beta_i + \beta_{i-1}}{\Delta t^2}
\label{eq:lateral_jerk_chain_rule}
\end{equation}
\subsubsection{Comfort Modeling via Jerk Constraints}
The computed jerk values in Equations~\eqref{eq:longi_jerk_final} and \eqref{eq:lateral_jerk_chain_rule} directly influence passenger comfort.
Longitudinal and lateral jerk constraints are defined as:

\begin{equation}
C_{\text{longi}} = \max_i \left| J_{\text{longi}}(i) \right| - \tau_{\text{longi}} 
\label{eq:longitudinal_jerk}
\end{equation}
\vspace{-0.5 cm}
\begin{equation}
C_{\text{lat}} = \max_i \left| J_{\text{lat}}(i) \right| - \tau_{\text{lat}} 
\label{eq:lateral_jerk}
\end{equation}

Where \( \tau_{\text{longi}}\)  and \( \tau_{\text{lat}}\) presented longitudinal and lateral comfort thresholds, respectively. We chose to fix these thresholds according to the investigation conducted in the study \cite{bae2019toward}, in which the jerk limit for a comfortable ride was found to fall within the range $ [\tau_{\min},\ \tau_{\max}] = [-0.9,\ 0.9]~\mathrm{m/s^3}$. 

\subsection{Efficiency Consideration}
\label{sec:Efficiency Consideration}
To achieve efficiency, we introduce a time-based indicator \textbf{Global Travel Time} noted \( \boldsymbol T_{global}\), defined as the elapsed time from the beginning to the end of a vehicle's maneuver.
In our notation, if \( t_0 \) denotes the initial timestamp at the starting location \( p_s \), and \( t_{\text{end}} \) denotes the final timestamp at which the vehicle reaches its goal destination \( p_g \), the total travel time is given by:
\begin{equation}
T_{\text{global}} = t_{\text{end}} - t_0
\label{ equ: Efficiency Consideration}
\end{equation}
Optimizing \( \boldsymbol  T_{global}\) prevents unnecessary delays and improves traffic efficiency.
\section{From Evaluation Metrics to a Unified Objective Function}
\label{sec:Global Objective Function}
Having established the safety, comfort, and efficiency criteria, we aggregate these components into a unified optimization framework that selects the best trajectory based on competing priorities. 
We define a reduced representation of the vehicle's state at the time \(t\) as:
\begin{equation}
    \Psi (t) =[s(t), v(t)]
    \label {equ:veh_state}
\end{equation}
where \( s(t) =[x(t), y(t)]\) $s(t)$ denotes the curvilinear abscissa along the reference trajectory, and $v(t)$ is the scalar velocity (i.e., the time derivative of $s(t)$).
The longitudinal dynamics are then given by:
\begin{equation}
\dot{s}(t) = v(t),
\quad
\dot{v}(t) = a(t),
\label{eq:vehicle_dynamics}
\end{equation}

where $a(t)$ is the longitudinal acceleration used as control input, constrained within the vehicle's physical limits:
\begin{equation}
a_{\min} \leq a(t) \leq a_{\max}, 
\quad 
0 \leq v(t) \leq v_{\max}.
\label{eq:acceleration_constraints}
\end{equation}
\(v_{\max}\) denotes the maximum speed limit allowed by traffic regulations.
To promote efficiency, the optimization framework seeks to minimize the total traversal time \( \boldsymbol  T_{global}\), defined in Equation \eqref{ equ: Efficiency Consideration}. However, a pure minimization of travel time may compromise safety and passenger comfort.
To address this, we formally structure the following constrained nonlinear problem: 
\begin{equation}
\begin{split}
Q &= \min_{\Psi (t)} \quad \omega_1 \cdot T_{\text{global}} + \omega_2 \cdot C_{\text{safe}} \\
\text{s.t.} & 
\left\{
\begin{aligned}
C_{\text{longi}} &\leq 0, \\
C_{\text{lat}} &\leq 0, \\
T_{\text{global}} &> 0, \\
a_{\min} &\leq \left\| a(t)\right\| \leq a_{\max}, \quad 0 \leq \left\| v(t)\right\|\leq v_{\max},\\
\end{aligned}
\right.
\end{split}
\label{eq:objective_function}
\end{equation}

where: \( C_\text{safe}\) denotes the safety criterion we designed in Equation \eqref{eq:safe_cost}. \( C_{longi}\) and \( C_{lat}\) are longitudinal and lateral jerk constraints as defined in Section \eqref{sec:Indicators_Confort}. 
In this formulation, $\omega_1$ and $\omega_2$ are weighting coefficients used to modulate the influence assigned to safety and efficiency, respectively, in the evaluation of the overall cost. These weights ensure dimensional consistency and allow potential scaling according to the respective magnitude of \( C_{\text{safe}}\) and $T_{\text{global}}$. Lastly, it provides flexibility to adjust the balance between safety and travel time to align with specific experimental requirements. 
The cost function \( Q\) balances time efficiency and ensures safety through \( T_{global}\) and \( C_{\text{safe}}\), respectively. The set of jerk constraints ensures passenger comfort.

Solving constrained optimization problems can be computationally demanding, especially when nonlinear and non-convex constraints are involved. To reduce the complexity of the problem expressed in Equation \eqref{eq:objective_function}, we incorporate the constraints into the main objective function via penalty terms. Thereby, transforming it into an optimization problem with only box-type constraints.
We add the term \( \psi(C_i)\) penalizing constraint violations and rewrite the former optimization problem as follows:
\begin{equation}
\begin{split}
Q &= \min_{\Psi (t)} \quad T_{\text{global}} + C_{\text{safe}} + \sum_i \psi_{(M_i,p_i)}(C_i) \\
\text{s.t.} & 
\left\{
\begin{aligned}
T_{\text{global}} &> 0, \\
a_{\min} \leq a(t) &\leq a_{\max}, \\
0 \leq v(t) &\leq v_{\max},\\
\end{aligned}
\right.
\end{split}
\label{eq:objective_function2}
\end{equation}

Here, the sum \( \sum_i \psi_{(M_i,p_i)}(C_i)\)  penalizes violations of the first two nonlinear inequality constraints (Jerk constraints) listed in Equation \eqref{eq:objective_function}. 

The penalty function $\psi_{(M,p)}$ enforces soft constraints by applying a penalty whenever a constraint is violated. Specifically, we define the penalty function $\psi_{(M,p)} : \mathbb{R} \rightarrow \mathbb{R}_+$ as a piecewise function:

\begin{equation}
\psi_{(M,p)}(x) =
\begin{cases}
0 & \text{if } x < 0 \\
\frac{1}{M}(e^{Mp x} - 1) & \text{if } x \geq 0
\end{cases}
\end{equation}

\begin{figure} [h!]
     \centering
     \begin{subfigure}[b]{0.4\textwidth}
         \centering
         \includegraphics[width=\textwidth]{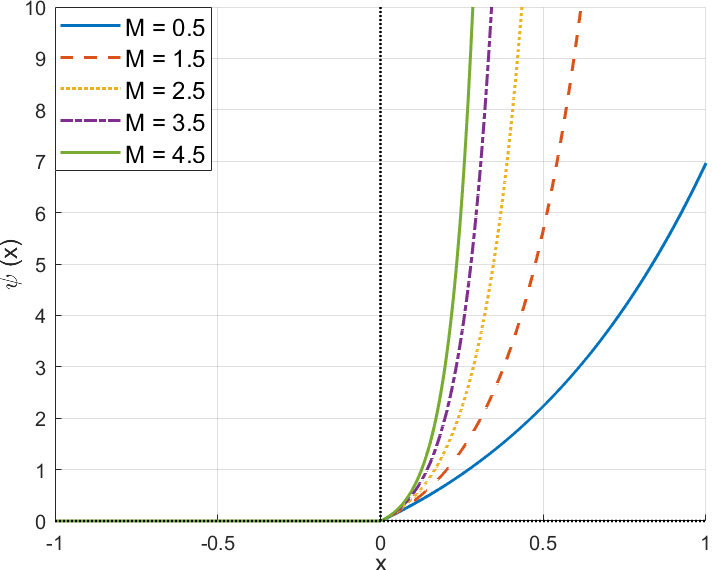}
         \caption{Influence of the parameter $M$ ($p$ fixed)}
         \label{fig:Mvar_Psi_func}
     \end{subfigure}
     \hfill
     \begin{subfigure}[b]{0.4\textwidth}
         \centering
         \includegraphics[width=\textwidth]{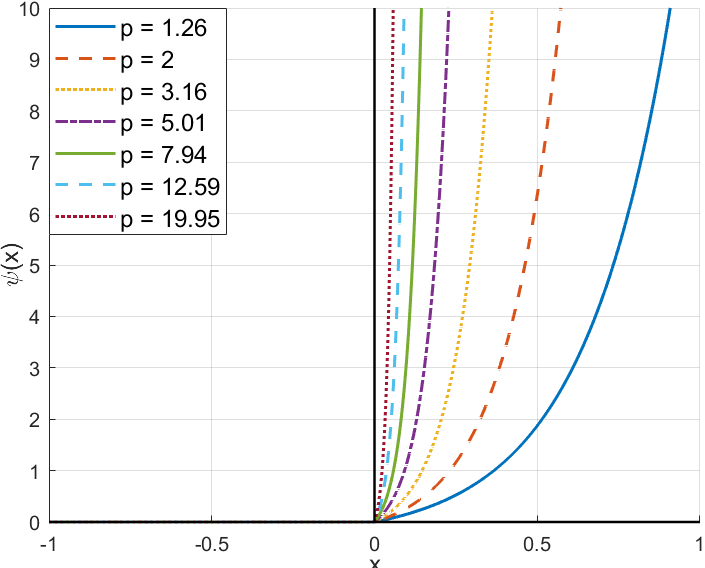}
         \caption{Influence of the slope $p$ ($M$ fixed)}
         \label{fig:Pvar_Psi_func}
     \end{subfigure}
        \caption{Illustration of the Penalty Function \(\psi(x) \) and the influence of parameters $M$ and $p$}
        \label{fig:PenaltyFunction}
\end{figure}

This function is continuous, convex over $\mathbb{R}$, and differentiable for $x \neq 0$.
The right-hand derivative at the origin is $\psi'(0^+) = p$, which allows fine control over the penalization severity for small violations. The parameter $M > 0$ controls the rate at which the penalty diverges: larger values of $M$ lead to a faster increase for $x > 0$, thereby enforcing stronger penalties for larger violations.
Figure \eqref{fig:PenaltyFunction} illustrates the influence of these parameters on the overall behavior of the function \(\psi (x)\).

By expressing the optimization problem in terms of penalties, this approach allows for separate tuning for minor and major violations, enabling more flexibility in the system’s response when needed during the optimization process. 

\section{Experimentation and simulation results}
\begin{figure} [H]
    \centering 
    \includegraphics[width=\linewidth]{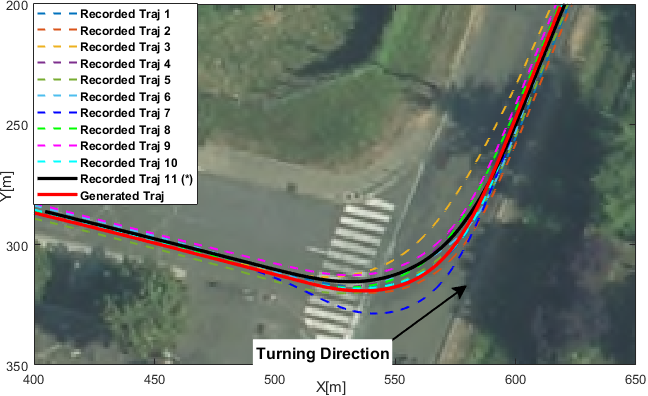}
    \caption{All.des Marronniers intersection: Comparison between the generated left-turn trajectory (solid red) and human-driven recorded trajectories. \small    The human recorded trajectories, depicted as colored dashed lines, were performed by a single driver, instructed to drive in the middle of the lane. Among the human recorded trajectories, the solid \textbf{black} one, labeled as $ (*) $, corresponds to the most comfortable trajectory, selected based on a minimum-jerk criterion, and was used as a reference for assessing the smoothness of our generated trajectory.
    \label{fig:humainVsauto}}
\end{figure}

The experimentation presented in this section has been performed using the experimental autonomous vehicle of VEDECOM Institute, on an unsignalized intersection located in All. des Marronniers, Versailles-Satory (78000), France. 
We selected 11 left-turn human driver trajectories at \href{https://maps.app.goo.gl/JtFLAhcfMSo9Xjup9}{All. des Marronniers intersection}. 
These recordings were compared to our generated trajectory, depicted as a solid red line in Figure \eqref{fig:humainVsauto}, which was produced under identical initial conditions and geometric layout parameters. 
The 11 recorded human trajectories were selected from a larger dataset to reflect diverse driving profiles observed in left-turn maneuvers at unsignalized intersections. Specifically, they capture conservative, balanced, and assertive behaviors. While the sample is limited, it is not intended for statistical inference, but rather to reflect representative behavioral archetypes.

\vspace {- 1 cm}
\subsection*{ $\bullet$ Comfort Evaluation} 
\begin{figure} [H]
    \centering 
    \includegraphics[width=\linewidth]{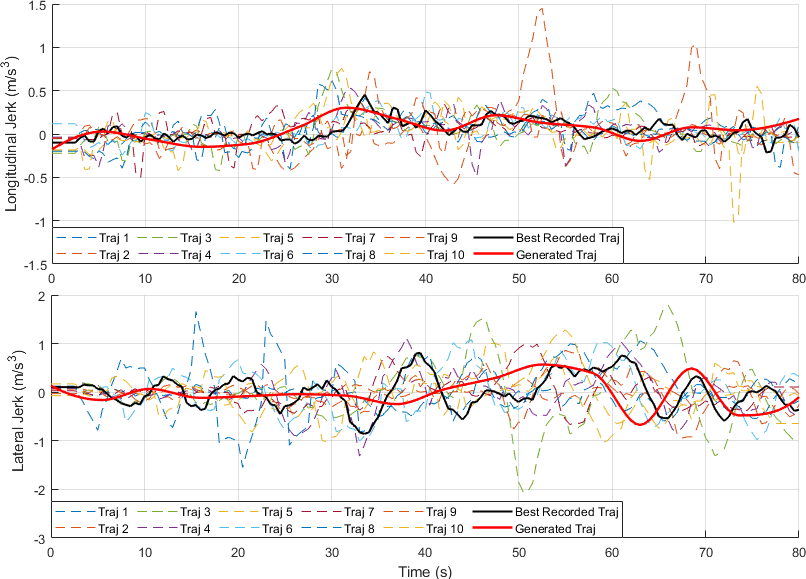}
    \caption{Longitudinal and Lateral Jerk Profiles for Left-Turn Trajectories at Allée des Marronniers Intersection}
    \label{fig:humainVsauto_Jerks}
\end{figure}

We report the longitudinal and lateral jerk profiles corresponding to the left-turn human trajectories, recorded at the Allée des Marronniers intersection, along with the jerk profile corresponding to our generated trajectory. The results are illustrated in Figure \eqref{fig:humainVsauto_Jerks}. Here, lower jerk values indicate smoother, more comfortable motion. The figures demonstrate that the generated trajectory maintains consistently lower jerk levels, thereby reducing abrupt accelerations and decelerations.


\subsection*{ $\bullet$ Safety Evaluation} 

This section presents a case study to illustrate the autonomous vehicle's behavior induced by the proposed optimization framework and its performance regarding safety considerations.
For this purpose, we present a scenario of a dynamic intersection crossing, involving interaction between an autonomous vehicle (Ego in red) approaching from the North-East and a human-driven vehicle (Opp in blue) approaching from the South-West. 
Here, the Ego vehicle decides its maneuver based on the outcomes of the optimization problem expressed previously in Equation \eqref{eq:objective_function2}. The experimental results are shown in Figures~\ref{fig:inter_met_Avant_stop}-~\ref{fig:inter_met_Apres_stop} and Figures~\ref{fig:interaction_case3}~(a)-(f). 

The aim of this analysis is to illustrate how the safety criterion \( C_{\text{safe}}\) introduced earlier in Equation \eqref{eq:safe_cost} automatically guides the planning towards a safe configuration, minimizing overlap situations between safety zones. 

\subsubsection*{ $\bullet$  \textbf{Full Stop by the Ego vehicle to Yield the Right of Way to the Opponent vehicle (Opp)}}

Figures~\ref{fig:interaction_case3}~(a)-(d) present the time evolution of 
the two agents as they converge towards the conflict zone with trajectories that are likely to collide. The ellipses around both vehicles represent the predicted occupancy at successive time steps. 

\begin{figure}[H]
\centering
\begin{subfigure}{\linewidth}
    \centering
    \includegraphics[width=0.7\linewidth]{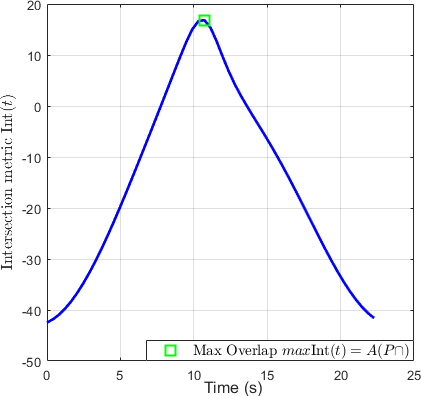}
\caption{Time evolution of the metric \( \text{Int}(t) \) without the decision making: The positive peak indicates a potential overlap between the safety zones}
        \label{fig:inter_met_Avant_stop}
\end{subfigure}

\vspace{2mm}

\begin{subfigure}{\linewidth}
    \centering
    \includegraphics[width=0.7\linewidth]{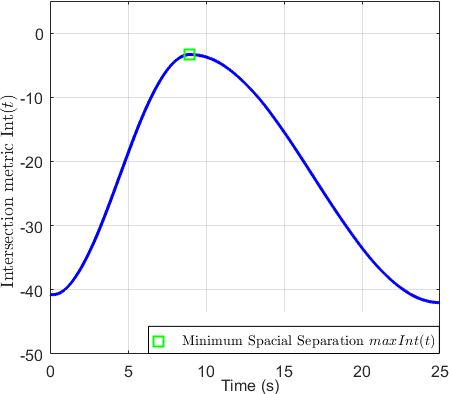}
    \caption{Time evolution of the metric \( \text{Int}(t) \) with the decision making: The curve remains strictly negative, ensuring safe separation}
    \label{fig:inter_met_Apres_stop}
\end{subfigure}
\caption{Time evolution of the metric \( \text{Int}(t) \): A sharp positive peak indicates a potential overlap between the safety zones, which was ultimately avoided due to the Ego’s full stop. The figure illustrates a transient violation that guides the algorithm toward safer trajectories.}
\label{fig:interaction_case3}
\end{figure}

\begin{figure*}[htp]
\centering

\begin{minipage}[t]{0.48\linewidth}
  \centering
  \includegraphics[width=\linewidth]{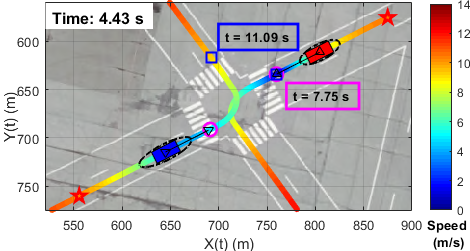}
  \small (a) Top View -- Initial Positions 
\end{minipage}
\hfill
\begin{minipage}[t]{0.48\linewidth}
  \centering
  \includegraphics[width=\linewidth]{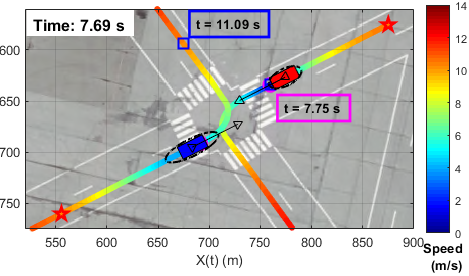}
  \small (b) Top View -- Beginning of Maneuver 
\end{minipage}

\vspace{0.7em}

\begin{minipage}[t]{0.48\linewidth}
  \centering
  \includegraphics[width=\linewidth]{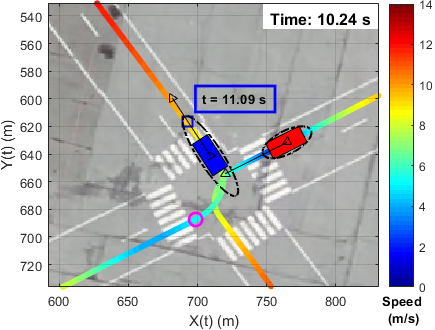}
  \small (c) Top View -- Conflict Zone Interaction 
\end{minipage}
\hfill
\begin{minipage}[t]{0.48\linewidth}
  \centering
  \includegraphics[width=\linewidth]{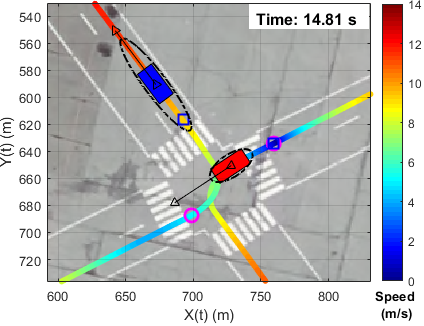}
  \small (d) Top View -- End of Maneuver 
\end{minipage}

\vspace{0.7em}

\begin{minipage}[t]{0.48\linewidth}
  \centering
  \includegraphics[width=\linewidth]{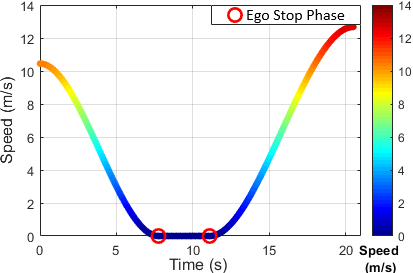}
  \small (e) Ego vehicle speed over experiment time 
\end{minipage}
\hfill
\begin{minipage}[t]{0.48\linewidth}
  \centering
  \includegraphics[width=\linewidth]{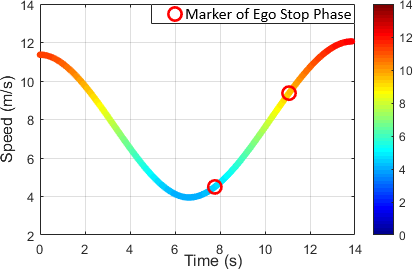}
  \small (f) Opp vehicle speed over experiment time 
\end{minipage}

\caption{ Full Stop by Ego to Yield Right of Way to Opp. The top-down sequence (a--d) captures the interaction stages, while plots (e--f) display velocity profiles.}
\label{fig:interaction_case3}
\end{figure*}


In this configuration, when no optimization strategy is used, the prediction of the trajectories of the two agents and their respective elliptical safety zones leads to a potential overlap. This overlap is characterized by a positive value of the metric \( \text{Int}(t) \), as shown by the positive peak of the curve in Figure \eqref{fig:inter_met_Avant_stop}. 
It should be emphasized that this plot does not correspond to a physically feasible situation. The peak positive value of the metric \( \text{Int}(t) \) indicates that the safety zones may overlap, and that the vehicles would have simultaneously occupied the same location, which is obviously impossible. 
In this figure, the worst case, marked by a square (in green), highlights the moment of maximum intrusion.
In the presented use case, the optimizer chooses a full stop for the ego vehicle, leading to \(\max _{t} \text{Int}(t) < 0 \) throughout the horizon.
Figure \eqref{fig:inter_met_Apres_stop} shows the final result after optimization. In this case, the indicator \( \text{Int}(t) \) remains negative over all times, which confirms that the ellipsoidal separation is fully maintained during the interaction.
Further explanation is given below: 

In Figures~\ref{fig:interaction_case3}~(b)-(c), we can see that the ego vehicle undergoes a deceleration phase ( marked here by the reduction of the major axis of the ellipse), followed by a complete stop before the intersection, allowing the opponent vehicle to pass through.
These observations are complemented by the vehicles' speed profiles represented in Figures~\ref{fig:interaction_case3}~(e)--(f).
Ego’s profile (Figure~\ref{fig:interaction_case3}~(e)) reveals a deceleration followed by a stationary phase of approximately \( 3.65\,\text{s} \), indicating a full stop to ensure safe passage for Opp. In contrast, Opp's speed profile (Figure~\ref{fig:interaction_case3}~(f)), shows a deceleration to negotiate the turn, followed by a smooth acceleration upon exiting the intersection, without coming to a complete stop.

The Ego vehicle's behavior emerges from the minimization of the global cost function \eqref{eq:objective_function2}. Specifically, the shaping function $ \beta (x) $  ( i.e.  \eqref{eq:beta_function}) imposes a sharp penalty for \( \text{Int}(t) > 0 \), discouraging safety ellipses overlap.  

The evaluation is performed during the optimization process to inform the algorithm that this configuration is infeasible and must be excluded for an alternative solution in which the safety ellipses remain disjoint over the entire time horizon. 
This strategy guarantees a maneuver that allows a trade-off between safety, comfort, and efficiency, as pointed by the objective function.

These results validate the geometric safety formulation within the global cost function as an effective guidance mechanism for the optimization process.   
The resulting behavior illustrates how a properly weighted cost formulation naturally induces safe and interpretable interaction strategies, without the need for explicit rule-based decision-making.

\section{CONCLUSIONS}

This paper introduced a set of metrics for assessing vehicle trajectory based on three complementary axes: safety, comfort, and efficiency. 
The safety criterion is addressed through a set of collision-aware geometric indicators. 
Comfort was modeled using jerk-based indicators to quantify ride quality from a passenger perspective.
Efficiency considerations are integrated through the minimization of the total travel time to ensure that the proposed solution does not lead to overly conservative behavior. 
Finally, these criteria were merged into a unified objective function, laying the basis for the optimization problem introduced in this paper as well.

This structure provides a modular yet unified framework: each criterion can be analyzed independently to evaluate a given characteristic of a trajectory or aggregated into a unified indicator for multi-objective evaluation, whether for trajectory generation or post-hoc analysis. 
In addition, the proposed criteria may serve as a benchmark or diagnostic tool to assess the quality of trajectories obtained through any method, including optimization, data-driven approaches, or human driving behavior.

However, autonomous vehicle navigation is inherently a multi-agent problem, where each vehicle’s actions influence, and are influenced by, those of other traffic participants. Future directions may investigate the use of the presented optimization problem in a decision-making process that accounts for interactions in multi-agent traffic environments.

 \bibliographystyle{IEEEtran}
 \bibliography{ifacconf}   

\end{document}